\documentclass[letterpaper, 10 pt, conference]{ieeeconf}  % Comment this line out if you need a4paper

\IEEEoverridecommandlockouts                       

\usepackage{amsmath,amssymb,amsfonts}
\usepackage[ruled,linesnumbered]{algorithm2e}
\usepackage{algorithmic}
\usepackage{graphicx}
\usepackage{textcomp}
\usepackage{verbatim}
\usepackage{threeparttable}
\usepackage{multirow}
\usepackage{xcolor}
\SetKwRepeat{Do}{do}{while}

\newcommand{\nop}[1]{}

\title{\LARGE \bf
Improving the Generalizability of Trajectory Prediction Models with Frenét-Based Domain Normalization}

\author{Luyao Ye, Zikang Zhou, and Jianping Wang% <-this % stops a space
\thanks{All authors are with the Department of Computer Science, City University of Hong Kong, Hong Kong SAR, and City University of Hong Kong Shenzhen Research Institute, Shenzhen, China. Emails: \{luyaoye2-c, zikanzhou2-c\}@my.cityu.edu.hk; jianwang@cityu.edu.hk. 
        }
\thanks{This work was partially supported by Hong Kong Research Grant Council under GRF 11200220, Science and Technology Innovation Committee Foundation of Shenzhen under Grant No. JCYJ20200109143223052. }
        }

\begin{document}

\maketitle
\thispagestyle{empty}
\pagestyle{empty}

%%%%%%%%%%%%%%%%%%%%%%%%%%%%%%%%%%%%%%%%%%%%%%%%%%%%%%%%%%%%%%%%%%%%%%%%%%%%%%%%
\begin{abstract}

Predicting the future trajectories of robots' nearby objects plays a pivotal role in applications such as autonomous driving. While learning-based trajectory prediction methods have achieved remarkable performance on public benchmarks, the generalization ability of these approaches remains questionable. The poor generalizability on unseen domains, a well-recognized defect of data-driven approaches, can potentially harm the real-world performance of trajectory prediction models. We are thus motivated to improve models’ generalization ability instead of merely pursuing high accuracy on average. Due to the lack of benchmarks for quantifying the generalization ability of trajectory predictors, we first construct a new benchmark called argoverse-shift, where the data distributions of domains are significantly different. Using this benchmark for evaluation, we identify that the domain shift problem seriously hinders the generalization of trajectory predictors since state-of-the-art approaches suffer from severe performance degradation when facing those out-of-distribution scenes. To enhance the robustness of models against domain shift problems, we propose a plug-and-play strategy for domain normalization in trajectory prediction. Our strategy utilizes the Frenét coordinate frame for modeling and can effectively narrow the domain gap of different scenes caused by the variety of road geometry and topology. Experiments show that our strategy noticeably boosts the prediction performance of the state-of-the-art in domains that were previously unseen to the models, thereby improving the generalization ability of data-driven trajectory prediction methods. The code can be found at \url{https://github.com/XIAOYEJIAYOU/Frenet-Strategy}.

\end{abstract}

%%%%%%%%%%%%%%%%%%%%%%%%%%%%%%%%%%%%%%%%%%%%%%%%%%%%%%%%%%%%%%%%%%%%%%%%%%%%%%%%
\section{INTRODUCTION}

The task of trajectory prediction is one of the indispensable components in safety-critical robotic applications, e.g., autonomous driving and robot obstacle avoidance. Given objects' past trajectories and the associated scene context, such as high-definition (HD) map, the goal of trajectory prediction is to predict objects' future movements and thereby enable safe motion planning of robots. Recent research in trajectory prediction has witnessed the huge success of deep learning. With their strong capability of fusing heterogeneous information in the scene, deep learning approaches have dominated the public benchmarks for trajectory prediction~\cite{zhan2019interaction, ettinger2021large, Chang_2019_CVPR, wilson2021argoverse}. However, whether these data-driven models can be generalized to out-of-distribution (OOD) scenes is still undetermined. 
\begin{figure}[htpb]
    \centering
    \includegraphics[width=\linewidth]{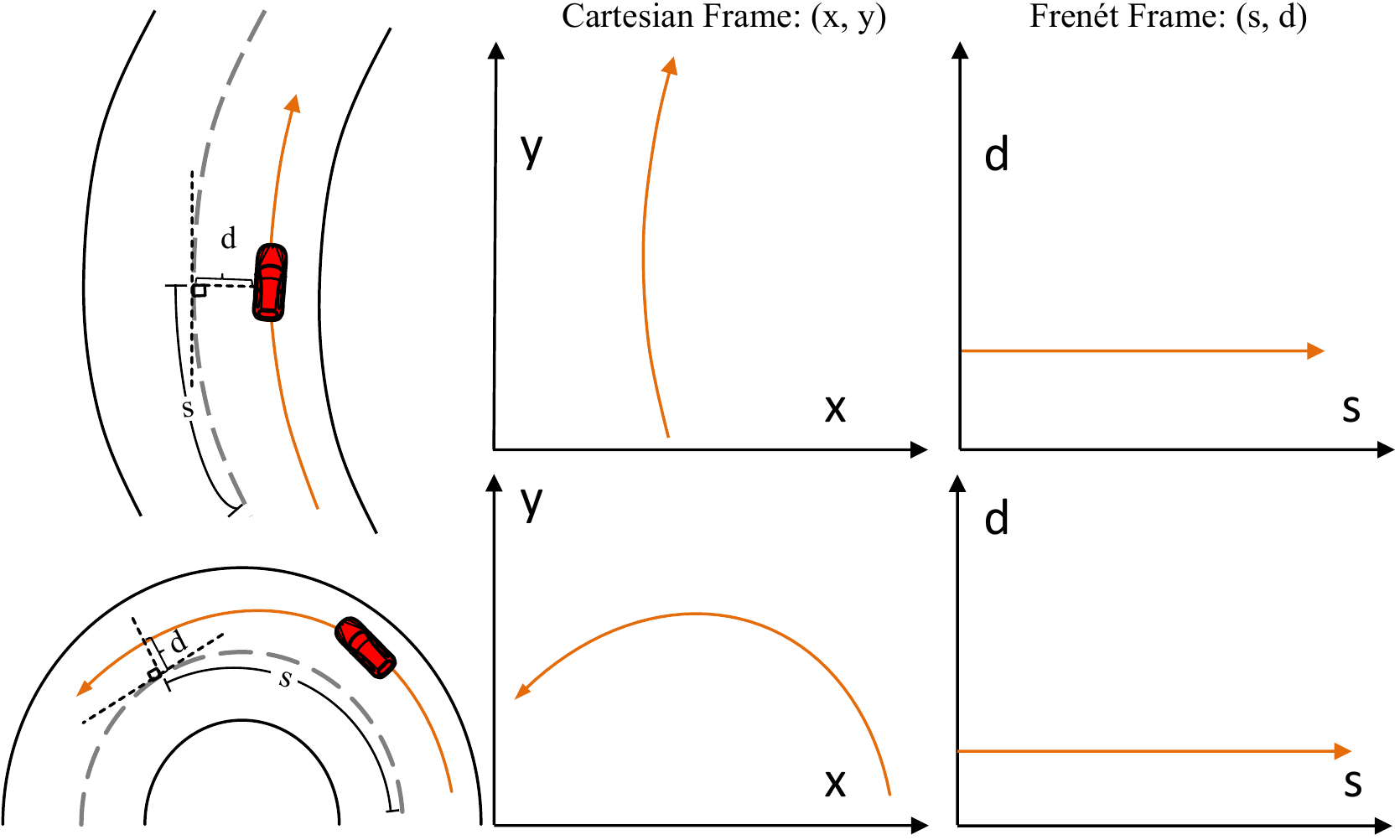}
    \caption{The solid black line represents the road boundary, the dotted gray line represents the centerline, and the orange arrow represents the vehicle trajectory.}
    \label{fig1}
\vspace{-0.5cm}
\end{figure}
As a well-known issue of learning-based approaches, the performance depends heavily on the distribution of training data and is prone to be affected by the distribution shift problem. Namely, these models usually have satisfactory accuracy in scenes frequently appearing in training data but may have trouble making correct predictions when facing those less frequent ones. For example, a trajectory predictor fully trained on the highway dataset can perform well on straight roads that were unseen before but is very likely to fail when tested on roundabouts due to the different data distributions on straight roads and roundabouts. Collecting sufficient data in all domains for training is not an affordable or feasible solution. If a model cannot make reliable predictions in unseen situations, catastrophic accidents may happen in the real world. For this reason, there is an urgent need to enhance the generalization ability of trajectory prediction models.
 
In daily traffic scenarios, traffic participants do not move in free space but need to obey traffic rules, e.g., driving in lanes or walking on the sidewalk. These traffic rules are mostly conveyed by the map. Therefore, many advanced trajectory prediction approaches~\cite{liang2020learning, gao2020vectornet, Zhou_2022_CVPR} focus on modeling the interactions between objects and HD maps to assist trajectory prediction. However, the geometry and topology of the map elements (e.g., the curvature of lanes) vary dramatically in different scenes, which brings the distribution shift problem and makes it difficult for the model to generalize in OOD scenes~\cite{bahari2022vehicle}. To address the above issues, we propose a domain normalization method, termed as Frenét+, for eliminating the difference among scenes by utilizing the Frenét frame~\cite{werling2010optimal}. This method explicitly moderates the distribution shift problem caused by the diversity of HD maps and enables the trajectory prediction models to focus more on domain-independent features (e.g., motion patterns of objects and social interactions among traffic participants) rather than overfitting the training data by memorizing the domain-specific map features. Specifically, we calculate the relative coordinates of the target object with respect to the centerline and use the relative coordinates for modeling. Here, we use the Frenét coordinate (i.e., the arc length and the perpendicular offset of the centerline) to represent the relative position of the target object. As shown in Fig.~\ref{fig1}, converting to the Frenét coordinate has a significant advantage of reducing the difference among the road shapes of different traffic scenes. We expect that existing trajectory prediction models combined with this domain normalization technique will be able to perform almost identically well on the seen and unseen domains.

To verify the existence of the domain shift problem and the effectiveness of our method, we first propose an automatic domain split schema based on a clustering algorithm to construct a new benchmark named argoverse-shift for model evaluation. After splitting different domains, we divide these domains into training set, seen validation set, and unseen validation set. Then, we evaluate several state-of-the-art models on this benchmark and observe that their performance on the unseen validation set is much worse than that on the seen validation set. After integrating with our domain normalization approach, the performance of these models on unseen domains is substantially improved.  

The key contributions are summarised as follows:
\begin{itemize}
\item We propose an automatic domain split schema and construct a new benchmark for evaluating the generalization ability of trajectory prediction models against the distribution shift problem.
\item We design validation experiments to explicitly quantify the generalizability of learning-based trajectory prediction models using our benchmark.
\item We introduce a plug-and-play strategy for domain normalization based on the Frenét frame to help the model recover from the distribution shift problems. Experiments show that this strategy noticeably enhances the prediction performance of state-of-the-art methods in the unseen domains.

\end{itemize}
    
\section{Related Work} 
\label{sec:relatedwork}

\subsection{Trajectory prediction}
Trajectory prediction is a classic problem of autonomous driving and has been widely studied in recent years. Early approaches to trajectory prediction are only based on historical trajectories of the ego vehicle and neighborhoods. While models are rapidly updating, using the LSTM network ~\cite{alahi2016social, altche2017lstm, deo2018convolutional}, Convolutional Neural Network ~\cite{nikhil2018convolutional}, Generative Adversarial Networks ~\cite{Gupta_2018_CVPR, kosaraju2019social, sadeghian2019sophie} and graph neural networks ~\cite{liang2020learning, mohamed2020social, ye2021gsan}, all these models neglect the influence of map information on trajectory prediction, so as to make it difficult to breakthrough in terms of prediction accuracy. Thanks to the development of High Definition Map and the release of new trajectory prediction datasets ~\cite{zhan2019interaction, Chang_2019_CVPR, ettinger2021large}, recent works focus on capturing scene representation from HD map in order to improve the performance. Some suggested using the image learning ability of CNN to represent traffic scenes on the map ~\cite{Hong_2019_CVPR, cui2019multimodal}. Another solution has rasterized map elements from the HD map as model inputs ~\cite{hong2019rules, chai2019multipath, cui2019multimodal, chou2020predicting, djuric2020uncertainty, gilles2021home}. While raster map representation is popular, this method was replaced by vectorized map data due to its efficiency. Vectorized methods ~\cite{liang2020learning, gao2020vectornet, Zhou_2022_CVPR} learn the relationships among entities in the scene as a set of vectorized entities with semantic and geometric features. Despite the good performance achieved by these methods, researchers have paid limited attention to the generalizability of trajectory prediction models to new domains. Our work proposes new solutions and evaluation benchmarks for this problem.

\begin{figure}[t]
    \centering
    \includegraphics[width=\linewidth]{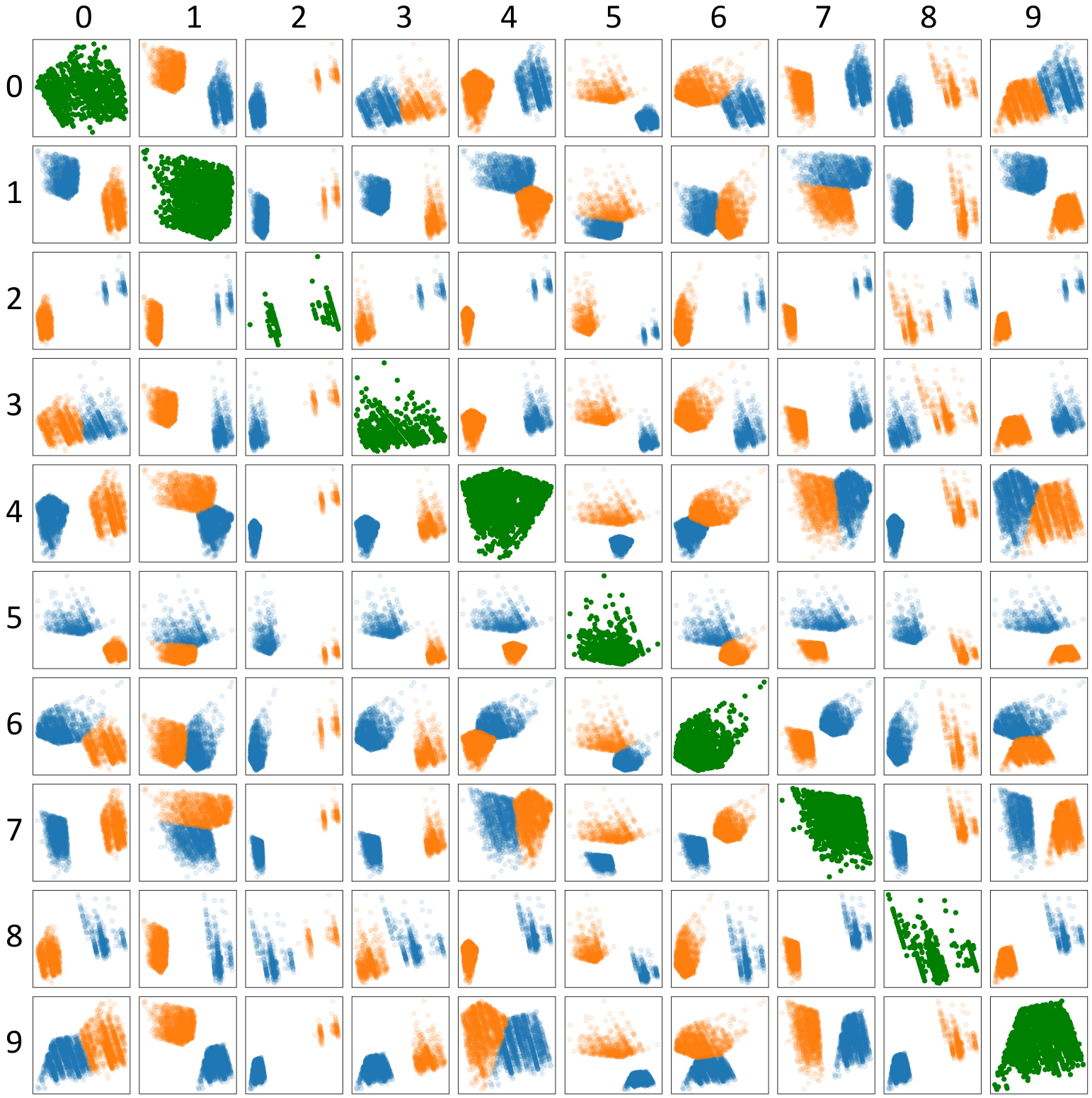}
    \caption{The 10 by 10 scatter plots matrix of domains overlaps. The cross of i-th row j-th column represents the overlap scatter between i-th domain and j-th domain. For each single scatter plot, the points in the same color belong to the same domain. The scatter plots on the principal diagonal show the points distribution of corresponding domains.}
    \label{fig:split}
\vspace{-0.5cm}
\end{figure}

\subsection{Coping with distribution shift in autonomous driving}
Prior work by Angelos et al. \cite{filos2020can} highlight the necessity of out-of-training-distribution scene detection in autonomous driving. They proposed a robust imitative planning method for detecting distribution shift and generating a safe plan. They also provided online supervision to efficiently query expert guidance for a safe course of action when extreme uncertainty. Another recent work demonstrated that sixty percent of existing scenes could be modified to make trajectory prediction models fail\cite{bahari2022vehicle}. They presented a scene generation model to provide richer and enough scenes for model training. To further evaluate the generalizability of trajectory prediction models, Thomas et al. \cite{https://doi.org/10.48550/arxiv.2205.07310} studied the performance of baseline models across four different datasets, and used the heap map to measure model uncertainty. However, no direct and effective method has been proposed yet to enhance the generalizability of trajectory prediction models. In contrast to the work mentioned above, we present a practical and effective solution to mitigate the effect of distribution shift in trajectory prediction, which does not require manual supervision, is model-independent, and can be directly used in current trajectory prediction models.    

\subsection{Frenét frame}
The Frenét frame based on the Frenet–Serret formulas \cite{frenet1852courbes} locally describes one point on a curve, a moving coordinate system determined by the curvature and the tangent line along the curve. In autonomous driving, many studies\cite{werling2010optimal,althoff2012line,verschueren2016time,fan2018baidu} have used the Frenét frame for safe and optimal motion planning. They assumed that the centerline was the ideal path along the free road. Therefore, they chose the centerline as the reference path and solved the motion planning problem in Frenét coordinates rather than Cartesian coordinates. The Frenét coordinates use the arc length of the centerline and the perpendicular offset to indicate the relative position of the points on the trajectory with respect to the centerline.   

Inspired by the Frenét frame method of motion planning, we apply Frenét frame to alleviate the impact of distribution shift in trajectory prediction. Finding the reference path in a complex scene and figuring out the projection of the given point are two critical challenges to transferring trajectory from Cartesian to Frenét frame. We proposed the Frenét+ strategy to solve the above problems.

\section{Preliminary Study}
\label{sec:ps}

In this section, we first present an automatic domain split schema since very few cross-domain datasets and benchmarks are available. Then, we demonstrate that the distribution shift problem exists in the state-of-the-art trajectory prediction models using our constructed benchmark.

\subsection{Domain split schema}
\label{subsec:dss}
In this work, we define the domain as a cluster of similar samples. For example, tracks sampled from the straight road and bend can be regarded as two domains.
Dividing the dataset into multiple domains needs to ensure that the data within each domain has unique features as much as possible. In other words, the gap between the data distribution of different domains should be large. In this way, the partitioned domains are more conducive to verifying the distribution shift problem. However, it is a challenge to find out the domain boundaries precisely. Designing the boundaries manually is not only a large workload but also easily influenced by individual subjectivity, leading to inconsistent split criteria. For this reason, we propose a clustering-based automatic domain split schema.

Specifically, we first perform feature extraction for each record in the dataset. We pre-define 21 features, such as the lane deflection angle, the differences of lane coordinates and the lane boundary. We extract these 21 features for $j$-th record to formulate a feature vector $d_j\in\mathbb{R}^{21}$. This 21-dimensional vectors are then downscaled to 2-dimensions $\hat{d}_j\in\mathbb{R}^{2}$ using the PCA algorithm \cite{krishna1999genetic}. 
Experiments show that the 2-dimensions feature is good enough to characterize a sample and also convenient for visualization.
Intuitively, these 2-dimensional vectors have a stronger characterization capability. The K-Means algorithm \cite{abdi2010principal} is then used to cluster these 2-dimensional vectors. In this way, each record is assigned with a cluster and we believe these clusters can be regarded as domains. Based on the clustering results, we retrace the corresponding data and match them with the corresponding domains. We further plot the distribution between each pair of divided domains to investigate the overlap among them. Fig. \ref{fig:split} shows no overlap between each domain pair and suggests that the split schema achieves the expectation.

\subsection{Argoverse-shift benchmark}

We construct a benchmark, referred to as argoverse-shift, to evaluate the generalization ability of trajectory prediction models against the distribution shift problem.

Following the domain split schema proposed in Section \ref{subsec:dss}, we partition Argoverse dataset\cite{Chang_2019_CVPR} into ten domains. We take the first seven domains as seen domains and the three left domains as unseen domains. The training set and validation set are sampled from all seen domains with a ratio of 8:2. All of the unseen domains are taken to formulate the test set. The detailed statistics are shown in Table \ref{tab:statistics}. On the one hand, the new cross-domain dataset, argoverse-shift, can be used to verify the existence of the distribution shift problem in the trajectory prediction task, and on the other hand, it can be used as a new benchmark to evaluate the trajectory prediction model's generalizability. 

\begin{table}[t]
\centering
\caption{Argoverse-Shift Dataset Statistics}
\label{tab:statistics}
\begin{tabular}{lccc}
\hline
\multicolumn{1}{c}{Subset} & Domain ID                    & Ratio (rel.) & Volume  \\ \hline
Train                      & 0, 1, 2, 3, 4, 5, 6          & 58.35\%      & 143,202 \\
Validation                 & 0, 1, 2, 3, 4, 5, 6          & 14.59\%      & 35,800  \\
Test                       & 7, 8, 9                      & 27.06\%      & 66,412  \\ \hline
All                        & 0, 1, 2, 3, 4, 5, 6, 7, 8, 9 & 100\%        & 245,414 \\ \hline
\end{tabular}
\vspace{-0.5cm}
\end{table}

To verify that current trajectory prediction models suffer from the distribution shift problem, we select several state-of-the-art models and observe whether their performances degrade when the train and test sets are taken from different domains. Experiment results show the performance of well-trained models decreases in the unseen domains. The analysis and implementation details for the domain shift verification experiment are described in Section \ref{sec:experiment}.

\section{Problem Definition}
\label{PD}

We take $X \in \mathcal{X}$ to denote the input sampled from the input space $\mathcal{X}$ and $Y \in \mathcal{Y}$ to denote the output from the output space $\mathcal{Y}$. As recent works included many features into consideration, e.g., neighbors' coordinates and HD map, trajectory predication is not necessarily a self-regression task, i.e., $\mathcal{X}\neq\mathcal{Y}$. 

We suppose that the data in a given dataset $\mathcal{S}$ can be divided into $M$ independent domains, i.e., $\mathcal{S} = \{\mathcal{D}^1,\cdots, \mathcal{D}^M\}$, where $\mathcal{D}^i=\{(X_j^i, Y_j^i)\}_{j=1}^{n_i}$ denotes the $i$-th domain. The distributions between each pair of domains are different. The first $K$ domains of $S$ are taken as the seen domains $S_{seen}=\{\mathcal{D}^1,\cdots, \mathcal{D}^K\}$ that models can access during the training, and the rest are taken as the unseen domains $S_{unseen}=\{\mathcal{D}^{K+1},\cdots, \mathcal{D}^M\}$ that are used to simulate the new scenes appeared on the road. The training set is sampled from seen domains, i.e., $S_{train}\in S_{seen}$, and the validation set used during training is also from seen domain and does not overlap with the training set, i.e., $S_{val}\in \{S_{seen} \backslash S_{train}$\}. The test set is sampled from the unseen domains $S_{test} \in S_{unseen}$.

The goal of domain generalization in trajectory prediction task is to learn a robust and generalizable prediction function $h: \mathcal{X} \rightarrow \mathcal{Y}$ from the seen domains to achieve a minimum prediction error on the unseen test domain \cite{wang2022generalizing}:
\begin{equation}
    \min_h \mathbb{E}_{(X, Y)\in S_{test}}[\mathcal{L}(h(X), Y)],
\end{equation}
where $\mathbb{E}$ is the expectation and $\mathcal{L}$ is the loss function.

\begin{table*}[t]
\centering
\caption{The qualitative results of 5 baseline models in seen and unseen domains, compared with Frenét+ models}
\label{tab:result}
\begin{tabular}{lllllll}
\hline
\multicolumn{1}{l|}{\multirow{2}{*}{Model}} & \multicolumn{3}{c|}{Seen Domains}                                                     & \multicolumn{3}{c}{Unseen Domains}                                               \\
\multicolumn{1}{l|}{}                       & \multicolumn{1}{c}{minADE $\downarrow$} & \multicolumn{1}{c}{minFDE $\downarrow$} & \multicolumn{1}{c|}{MR $\downarrow$}     & \multicolumn{1}{c}{minADE $\downarrow$} & \multicolumn{1}{c}{minFDE $\downarrow$} & \multicolumn{1}{c}{MR $\downarrow$} \\ \hline
\multicolumn{1}{l|}{NN + MAP}               & 0.6342                            &1.3887             & \multicolumn{1}{l|}{0.1515}   & 1.9689 (+210.45\%)        & 3.7502 (+170.05\%)                & 0.5501 (+263.10\%)                        \\
\multicolumn{1}{l|}{LSTM ED + MAP}          & 2.0870                     & 4.4180                            & \multicolumn{1}{l|}{0.6485} & 2.2622 (+8.39\%)           & 4.7286 (+7.03\%)           & 0.6830 (+5.32\%)       \\
\multicolumn{1}{l|}{WIMP}                   & 0.7507                     & 1.1189                     & \multicolumn{1}{l|}{0.1092} & 0.8311 (+10.71\%)          & 1.2525 (+11.94\%)          & 0.1364 (+24.91\%)      \\
\multicolumn{1}{l|}{LaneGCN}                & 0.7152                     & 1.0974                     & \multicolumn{1}{l|}{0.1065} & 0.7653 (+7.01\%)           & 1.1554 (+5.29\%)           & 0.1148 (+7.79\%)       \\
\multicolumn{1}{l|}{HiVT}                   & 0.7642                     & 1.2081                     & \multicolumn{1}{l|}{0.1263} & 0.8595 (+12.47\%)          & 1.3836 (+14.53\%)          & 0.1505 (+19.16\%)      \\ \hline
\multicolumn{1}{l|}{Frenét+ NN + MAP}      & 0.8284                    & 1.7193                           & \multicolumn{1}{l|}{0.2114}       & 0.9984 (+20.52\%)                   & 2.0223 (+17.62\%)                            & 0.2625 (+24.17\%)                        \\
\multicolumn{1}{l|}{Frenét+ LSTM ED +MAP}  & 2.0918                     & 4.4296                           & \multicolumn{1}{l|}{0.6467} &2.1058 (+0.67\%)            & 4.4537 (+0.54\%)           & 0.6498 (+0.48\%)       \\
\multicolumn{1}{l|}{Frenét+ WIMP}          & 0.7596                     & 1.1263                     & \multicolumn{1}{l|}{0.1167} & 0.7718 (+1.61\%)           & 1.1475 (+1.88\%)           & 0.1183 (+1.37\%)       \\
\multicolumn{1}{l|}{Frenét+ LaneGCN}       & 0.7218                     & 1.1039                     & \multicolumn{1}{l|}{0.1064} & 0.7330 (+1.55\%)           & 1.1215 (+1.59\%)           & 0.1091 (+2.54\%)       \\
\multicolumn{1}{l|}{Frenét+ HiVT}          & 0.7756                     & 1.2370                     & \multicolumn{1}{l|}{0.1344} & 0.7882 (+1.62\%)           & 1.2602 (+1.88\%)           & 0.1402 (+4.32\%)       \\ \hline
                                            &                            &                            &                             &                            &                            &                       
\end{tabular}
\vspace{-0.6cm}
\end{table*}

\section{Frenét+ Strategy}
\label{method}

In this section, we first show the strategy to select the reference path. Then we illustrate the pipeline that transfers a trajectory point from the Cartesian frame to the Frenét frame based on the selected reference path. Finally, we provide the solution to determine the projections for those trajectory points located in the non-differential areas.

\subsection{The selection of the reference path}

\begin{figure}[t]
    \centering
    \includegraphics[width=\linewidth]{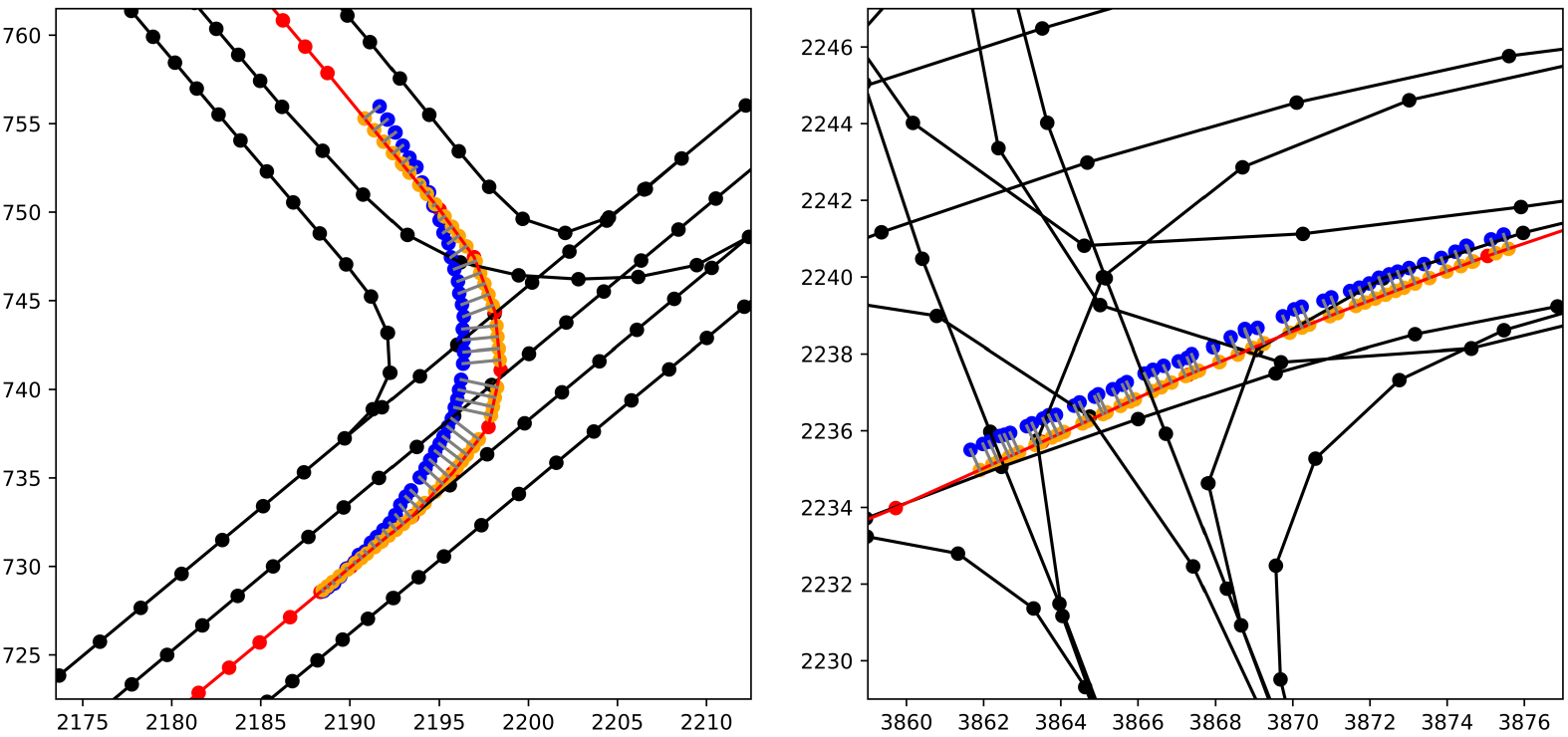}
    \caption{Trajectories with their projections. The solid black lines with points are centerlines that appeared in this scene. The centerline in red is selected as the reference path. The blue points are the trajectory points of the vehicle. The orange points are the projections on the reference path. The gray lines connect the trajectory points with their corresponding projections.}
    \label{fig:proj}
\vspace{-0.5cm}
\end{figure}

In complex scenes, such as the intersection, there are multiple centerlines, as shown in Fig. \ref{fig:proj}, which increases the difficulty of finding the appropriate reference path. To solve this problem, we proposed a method to determine a proper centerline as the reference path based on the vehicle's historical trajectory. Specifically, we first calculate the Euclidean distance between the vehicle's historical trajectory and each centerline and take the reciprocal of the distance as the similarity. 

\begin{equation}
    S{_1}^j = [\frac{1}{n}\sum_{t=1}^n{\|\mathrm{x}_t - \widetilde{\mathrm{x}}_t^j\|_2}]^{-1},
\end{equation}

where $\mathrm{x}_t\in \mathbb{R}^2$ and $\widetilde{\mathrm{x}}_t \in \mathbb{R}^2$ are the coordinates of the trajectory and corresponding projection on the centerline $j$ at the time step $t$. Intuitively, the closer centerline would be a better choice of the reference path.  

Next, we compare the shape similarity between centerlines and historical trajectory. Specifically, we translate candidate centerlines towards the vehicle and then calculate the Euclidean distance. The translation direction and length are defined by the current location of the vehicle with its projections on centerlines:

\begin{equation}
    \Delta \mathrm{x}^j  = \mathrm{x}_T - \widetilde{\mathrm{x}}_T^j,
\end{equation}
where $T$ indicates the current time step. Then the shape similarity is defined as:

\begin{equation}
    S{_2}^j = [\frac{1}{n}\sum_{t=1}^n{\|\mathrm{x}_t - (\widetilde{\mathrm{x}}_t^j + \Delta \mathrm{x}^j)\|_2}]^{-1}.
\end{equation}

After that, we take the average of the $S{_1}^j$ and $S{_2}^j$. Finally, we choose the centerline $j^*$ with the largest value as the reference line. 

\begin{equation}
    j^* = arg\max_{j} (S{_1}^j + S{_2}^j).
\end{equation}

\subsection{Coordinate transfer}
Suppose that a reference path is composed of $m$ segments and saved as a list of coordinate points following the order of the direction of the trajectory, i.e., $[p^1, \cdots, p^{m+1}]$. The projection of the vehicle on the reference path at the time step $t$ falls on the $J$-th segment, i.e., the segment bounded with the points $p^J$ and $p^{J+1}$. Based on these two points, we can easily calculate the expression of the line where the $J$-th segment is located, i.e., $y = k^Jx+b^J$. Then the projection of the trajectory point $\mathrm{x}_t=(x_t, y_t)$ at the time step $t$ can be represented as:
\begin{eqnarray}
    \widetilde{\mathrm{x}}_t&=&(\widetilde{x}_t, \widetilde{y}_t), \nonumber \\
    \text{where},\quad\widetilde{x}_t&=& \frac{k^J(y_t-b^J)+x_t}{(k^J)^2+1}, \nonumber \\
    \widetilde{y}_t&=&  k^J\widetilde{x}_t+b^J .
\end{eqnarray}

The offset of the trajectory point about the reference path, i.e., the $d$ coordinate in Frenét Frame, can be derived from the distance between the trajectory point and the projection:
\begin{eqnarray}
    d_t&=&I_t \cdot\|\widetilde{\mathrm{x}}_t-\mathrm{x}_t\|_2 , \nonumber \\
    \quad I_t&=&
    \begin{cases}
        \begin{array}{lr}
             -1,\quad\text{if}\quad\overrightarrow{p^{J}p^{J+1}}\cdot\overrightarrow{p^{J}x_t}<0 , &  \\
             \quad1,\quad\text{if}\quad\overrightarrow{p^{J}p^{J+1}}\cdot\overrightarrow{p^{J}x_t} \geq 0 . & 
        \end{array}
    \end{cases}
\end{eqnarray}
The $I_t$ is the indicator which indicates the side of the trajectory point relative to the reference path in the right-hand system. We initially set the first point of the reference path as the starting point. Then the arc length $s$ of the trajectory point in the Frenét frame can be expressed as the sum of some segments:
\begin{equation}
    s_t = \sum_{j=2}^J{\|p^j - p^{j-1}\|} + \|\widetilde{\mathrm{x}}_t - p^J\| .
\end{equation}
We take the difference between $s_t$ and $s_1$ as the final arc length, i.e., $s_t \xleftarrow[]{}(s_t - s_1)$. In this way, the projection of the first trajectory point is set as the starting point. Therefore, the arc length of the first trajectory point $s_1$ in the Frenét frame is 0.

So far, we present the whole process to convert a trajectory point from the Cartesian coordinate $(x_t, y_t)$ to the Frenét coordinate $(s_t, d_t)$. This process is unrelated to model design and can be easily adapted to most current trajectory prediction models. It can be set between the data pipe and the model, i.e., the coordinate transformation is performed before feeding trajectory into the model.

\subsection{Projections on the non-differentiable area}

Another challenge is to find the correct projections on the non-differentiable area of the reference line. From the engineering perspective, centerlines are often stored as a list of coordinate points. Connecting these points in order gives a series of line segments. In this case, the centerline is not a smooth curve but a polyline. Therefore, not all points have projections on the reference path. 

\begin{figure}[b]
    \centering
    \includegraphics[width=3in]{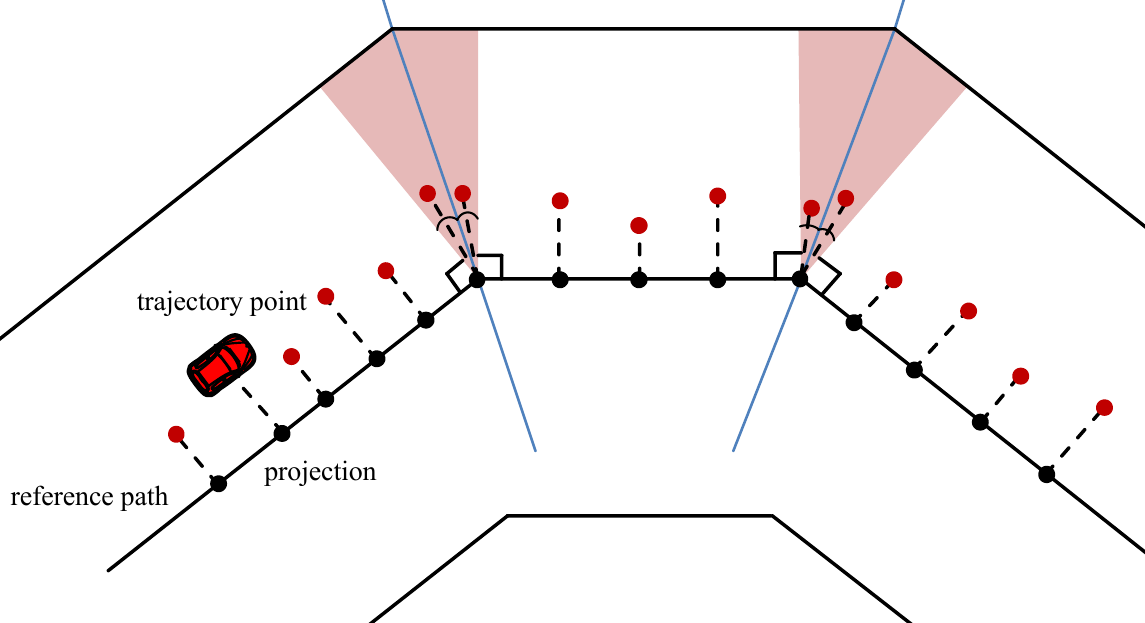}
    \caption{Find the correct projections on the non-differentiable area. The red points are trajectory points; the black points are the projections of the trajectory points on the reference path, and the blue line is the angle bisector.}
    \label{fig:proj2}
\end{figure}

As shown in Fig. \ref{fig:proj2}, we cannot find a projection for points in the red area since it is non-differential at joint points. We represent the red area in Fig. \ref{fig:proj2} at the joint $(x^*, y^*)$ as:
\begin{equation}
    \hat{p} = \{(x,y)|-\frac{1}{k_1}<\frac{y-y^*}{x-x^*}<-\frac{1}{k_2}\},
\end{equation}
where $k_1 > k_2$ are slopes of segments that intersect at the joint $(x^*, y^*)$.

We artificially set the projections of the trajectory points $(x,y)\in \hat{p}$ to the nearest endpoints on the reference line. Moreover, we set the angle bisector as the vertical direction at the joint point:
\begin{equation}
    y = \frac{1}{2}(\frac{1}{k_1}+\frac{1}{k_2})(x^*-x) +y^* .
\end{equation}

In this way, multiple points in the area can be converted to the same Frenét coordinate. It introduces extra errors when converting them back to the Cartesian coordinate frame. However, we experimentally verified that the error is less than $10^{-4}$ meters on average, which has a negligible effect on the final results.

Fig. \ref{fig:proj} shows that our Frenét+ strategy can find the proper reference path from a complex scene and get the correct projections on the non-differentiable centerlines.

\section{Experiments}
\label{sec:experiment}

In this section, we conduct comparative experiments by combining Frenét+ strategy with baseline models to demonstrate the effectiveness of our proposed strategy. We also perform result analysis and visualize several prediction results to give an insight into domain shift problem on trajectory prediction.

\subsection{Experimental settings}

The experiments are conducted on the argoverse-shift dataset. Following the same setting as Argoverse motion forecasting challenge \cite{Chang_2019_CVPR}, we require models to predict the position of the agents in the future 3 seconds, given the initial 2 seconds observations.

\textbf{Baselines:} We take NN+map \cite{Chang_2019_CVPR}, LSTM ED+map \cite{Chang_2019_CVPR}, WIMP \cite{khandelwal2020if}, LaneGCN \cite{liang2020learning}, HiVT \cite{Zhou_2022_CVPR}, the five representative state-of-the-art trajectory prediction models, as baselines.

\textbf{Metrics:} We employ three standard metrics for trajectory prediction, including the Minimum Average Displacement Error (minADE), Minimum Final Displacement Error (minFDE) and Miss Rate (MR). 
% The Average Displacement Error is defined as the average $l2$ distance in meters between the predicted trajectory and the ground truth over all predicted future time steps, while the Final Displacement Error measures the $l2$ distance in meters between the predicted and ground truth coordinates at the last time step. 
Models predict six trajectories, and we report the best result with minimum errors. 
% The miss rate refers to the ratio of predictions whose final location is more than 2.0 meters away from the ground truth. 

We train each baseline on the training set with the best practice of the hyperparameters reported in the original paper and select the best parameter group via the validation set. After training, we separately evaluate the model on the test set, i.e., the unseen domains, and the validation set, i.e., the seen domains, for comparison. We then apply the Frenét+ strategy to each model. We train and re-evaluate them, following the same process. The comparison results are reported in Table \ref{tab:result}.

\subsection{Result analysis}

Table \ref{tab:result} shows the performance of the five models on the seen domain and unseen domain, respectively, and their performance after combining the Frenét+ strategy. A smaller value means better performance. Due to changes in the dataset volume and split scheme, the results of baseline models are slightly different from the performance reported in the original paper. The value in parentheses is the performance deterioration on unseen domains relative to the performance on seen domains. 

\begin{figure}[t]
    \centering
    \includegraphics[width=\linewidth]{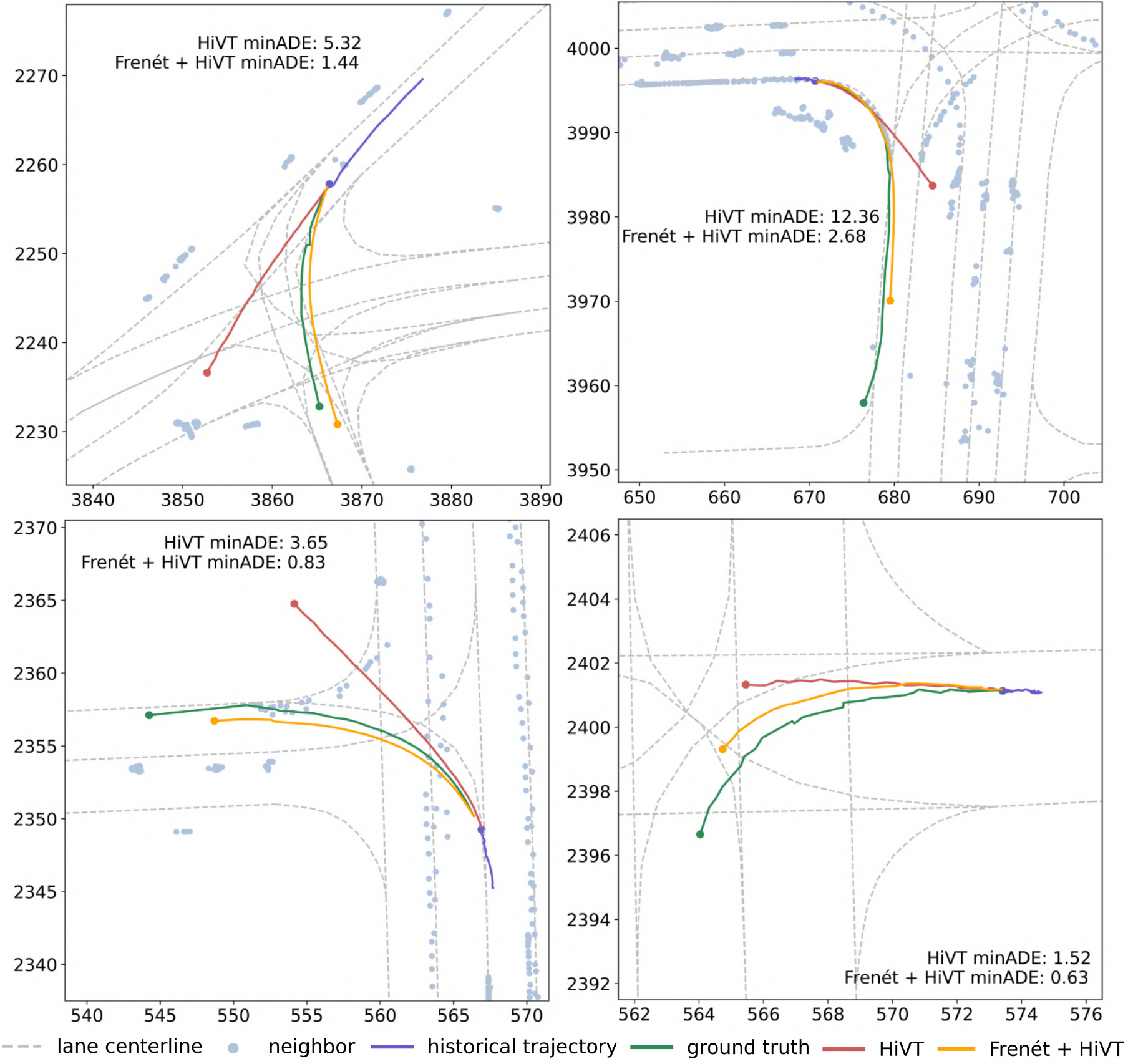}
    \caption{Visualized trajectory prediction results of HiVT and Frenét+ HiVT. The green line is the ground truth, the red line is the predicted trajectories by HiVT and the yellow line is predicted by Frenét+ HiVT.}
    \label{case}
\vspace{-0.5cm}
\end{figure}

\textbf{Domains shift problem verification:} From the top half part of Table \ref{tab:result}, the performance of well-trained models decreased on the unseen domains in terms of all evaluation metrics. The result illustrates that the domain shift problem exists exactly in current models. For those naive models, e.g., NN + MAP, this problem has a more serious negative impact. By comparison, LaneGCN is more robust against domain shift. It has the slightest drop in performance on the unseen domain. Because LaneGCN takes into account a large amount of information about the relative positions of centerlines, their predictions are less susceptible to topographic changes.

\textbf{Frenét+ effectiveness evaluation:} By comparing the upper and lower parts of Table \ref{tab:result}, we find that our Frenét+ strategy gives a significant improvement across all baselines on unseen domains in terms of all metrics. Though it is still not as good as the results on seen domains, the difference is not significant. Frenét+ brings more improvements to those simple models, like NN + MAP, since strong models are more robust in design.

We also find a slight decrease in the performance of the model with Frenét+, compared with the original one, on seen domains, e.g., the MinADE of HiVT increased from 0.7642 to 0.7756. We believe this is the necessary cost to improve generalization ability of baseline models, since we make the model focuses on general features rather than domain-specific details during training in order to obtain better generalizability.

\subsection{Visualization of prediction results and error distribution}

In Fig. \ref{case}, we visualize several prediction results of HiVT compared with Frenét+ HiVT. In the four scenes, the predicted trajectories by HiVT deviated from the ground truth and even crossed three lanes in the third scene, which is abnormal driving behavior. With the help of Frenét+ strategy, the predicted trajectory is corrected by the reference path and closer to the ground truth. It shows that the Frenét+ strategy provides a strong reference for models and has a significant effect on normalized abnormal prediction.

\begin{figure}[t]
    \centering
    \includegraphics[width=\linewidth]{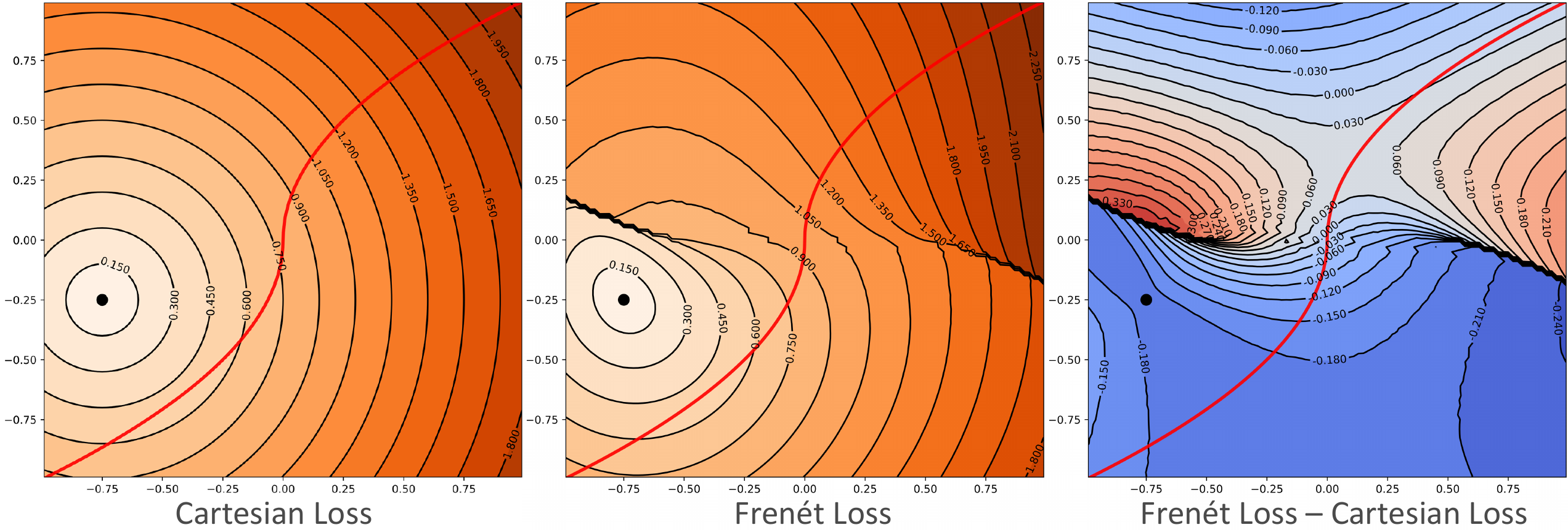}
    \caption{Contour maps of error distribution in Frenét frame, compared with Cartesian frame. The black point indicates the true trajectory point, i.e., the point with 0 error. The red curve is the reference path. The lighter the color, the less the error, and vice versa.}
    \label{loss}
\vspace{-0.5cm}
\end{figure}

As for the error distribution in the Frenét frame and Cartesian frame, it is obvious that they are exactly the same when the reference path is a straight line. We show a case with a curve reference path in Fig. \ref{loss}. Compared with Cartesian frame, the error distribution in Frenét frame is not a standard circle. When the projection falls on the part of the reference path with a large gradient, the changes of the loss are more dramatic. The third picture shows the difference between loss in the Frenét frame and Cartesian frame. 
% The red area indicates that the loss is larger in the Frenét frame, while the blue area indicates the loss is larger in the Cartesian frame. 
Though the error distributions are not the same in these two systems, the difference is limited. Hence, we claim that training models in the Frenét frame will not introduce significant optimization gaps, and the optimization objectives, i.e., the 0-error point, are the same. 

\section{CONCLUSIONS}
\label{conclude}

In this paper, we have introduced a new benchmark called argoverse-shift to verify that the domain shift problem does exist in data-driven trajectory prediction models. Then, we have proposed a Frenét-based strategy, Frenét+, to enhance the robustness of models against domain shift. Our approach can diminish the variation of trajectory coordinates across domains by exploiting the local coordinates of trajectory waypoints relative to the lane centerlines. Experiments and visualization results show that the Frenét+ strategy significantly mitigates the domain shift problem and makes state-of-the-art models generalize better on unseen domains. In the near future, we plan to explore the domain shift problem on more datasets, such as Waymo Open Dataset \cite{ettinger2021large} and NuScenes Dataset \cite{caesar2020nuscenes}.
                        
\clearpage
\bibliographystyle{IEEEtran}
\bibliography{ref}

\end{document}